# DEVELOPMENT OF ONTOLOGY-BASED INTELLIGENT SYSTEM FOR SOFTWARE TESTING

A. Anandaraj[1] P. Kalaivani[2] V. Rameshkumar[3]

[1&2]Department of Computer Science and Engineering, Narasu's Sarathy Institute of Technology, Salem, Tamilnadu.

[3]Department of Information Technology, Narasu's Sarathy Institute of Technology, Salem, Tamilnadu.

anandarajme@gmail.com, kalaipadman@gmail.com, vrameshbtech@gmail.com

***Abstract***

Software testing is a prime factor in software industry. Besides knowing the importance of testing, only limited time is allocated for teaching it. It will be more efficient if testing is taught simultaneously with programming foundations. This integrated learning of testing techniques and programming allows the programmers to perform in a better way and this leads to the improvement of the performance of the industry progress. In this paper, a technique named ontology is introduced, it first defines the various testing process in hierarchy and define relationships among them, to share and reuse the knowledge that is captured, secondly metadata is created by natural language processing and finally, the application use ontologies to support test management, it act as knowledge base for multiple environment with the integrated teaching of programming foundation and testing concepts.

## 1. INTRODUCTION

Software defects exist in almost all types of software with moderate size. To overcome these defects software testing is carried out. Software testing is one of the key assets of a software engineer to protect software from bugs but it is difficult to learn or teach without proper guidance.

During the learning process, testing is taught only at the last part and this does not allow software professionals to perform testing in a better way. Earlier mastering of testing concepts and techniques would: (1) improve the reasoning about the program (and its solution), leading to better high quality products; and (2) induce and facilitate the use of testing throughout the software development process, leading to a better high quality process, in contrast to the current practices [1].

One way to improve the efficiency of testing in the programmers is to introduce an ontological concept which integrates the teaching of programming and testing in a combined manner during the training period in industry.

Programming foundations is not an easy subject to be taught - many students have difficulties in understanding the concepts of programming [2] and have a wrong view about the programming activity [3]. Thus, it is a biggest challenge for the trainers to teach them in an efficient and effective way.

Testing requires the learners to know the behavior of their programs, such activity could be explored to help them understand the abstract concepts of programming and develop the expected skills [3].

The goal of this paper is to help software industry to get their fresher trained with programming concepts and testing simultaneously with the help of ontology. This paper describes the classification of programming and testing using protégé. This also defines the classes, properties and features such as reasoners to check semantic consistency. Finally, the paper shows the graphical view of the classes, generated OWL schemas and XML scripts which is used to link existing web resources into the semantic web.

## 2. DEVELOPMENT OF ONTOLOGIES USING PROTÉGÉ

OWL ontology is described as a network of classes, properties and individuals. Classes define names of the relevant domain concepts and their logical characteristics. Properties (also called as roles, attributes or slots) that defines relationships between classes, and allow assigning primitive values to instances. Individuals are instances of the classes with specific values for the properties [4].

The semantic web can be regarded as a network of ontologies and other web resources [4]. OWL ontology concepts can have references to concepts in other ontologies. The basic mechanism for this capability is ontology import (i.e., ontology can import resources from existing ontologies and create instances or specializations of their classes)

### *2.1 Design of Ontology*

The steps followed to build ontology are explained below:

#### *2.1.1 Determine Domain and Scope of Ontology*

The main goal of this work is to provide a framework for programming foundations and testing simultaneously. The objective is to use this ontology for integrated teaching of programming foundations and testing in software industry.

#### *2.1.2 Defining Concepts in the Ontology*

The terminologies that relates to objective are listed to create ontology. For example, important terms related to this are testing phase, testing techniques, oop paradigm, control flow statements etc. Figure 5 explains the testing and programming concepts.

#### *2.1.3 Create a Class Hierarchy*

The terminologies form the classes in the ontology. For example, encapsulation, error based technique, unit testing, and integration testing forms classes in OWL ontology and are represented in figure 2.

#### *2.1.4 Defining Properties and Constraints*

There are two types of properties viz., object properties and datatype properties. Object properties links object to an object. Datatype properties links objects to the XML schemas and are depicted in figure 3.

## 3. DEVELOPMENT OF ONTOLOGIES FOR SOFTWARE TESTING

The development of intelligent ontology based learning system for Software Testing is briefly illustrated in following steps.

### *3.1 Description of Classes*

The important view in the Protégé OWL plugin is the OWL classes. Classes describe concepts in the domain. This tab displays the tree of the ontology’s classes on the left. The upper region of the class is shown in a form in the center. This form allows users to edit class metadata such as name, comments, and labels, in multiple languages. The widget in the right area of the form allows users to assign values for properties and description to a class.

Annotation properties can be used to add information (metadata-data about data) to classes. Ontologies can define their own annotation properties or reuse existing ones such as those from the Dublin core ontology [4]. In contrast to other properties, annotation properties do not have any formal meaning for external OWL components like reasoners, but they are an extremely important vehicle for maintaining project information.

In this paper many classes and sub classes have been created under the field of testing and programming but due to lack of space only some of the classes are described elaborately.

Here the class testing techniques has sub classes error based technique, functional technique and structural technique. The class error based technique is further divided into error seeding, error guessing and mutation analysis. The classes are represented in figure 1.

The editing of classes is carried out using the classes tab shown in Figure 2. The initial class hierarchy tree view should resemble the picture shown in Figure 2. The empty ontology contains one class called Thing. The class Thing is the class that represents the set containing all individuals. Because of this all classes are subclasses of Thing [5]. To add a class, the classes tab is selected, add subclass button is pressed. This creates a new class as a subclass of the selected class Thing.

### *3.2 Creation of Properties*

The properties widget of the OWL classes tab allows users to view and create relationships between classes. It provides access to those properties that could be used by the instances of the current class. The characteristics of a property are edited through the form shown in Figure 4. This form provides a metadata area in the upper part, displaying the property's name, annotations and so on, similar to the presentation in the class form.

There are two main types of properties viz. Object properties and Datatype properties. Object properties are relation between two individuals. Object properties link an individual to an individual whereas datatype property links an individual to an XML Schema Datatype value or an RDF literal (i.e. they describe relationship between an individual and data values). OWL also has another property named annotation property, which is used to add information (i.e. metadata - data about data) to classes, individuals and object/ datatype properties.

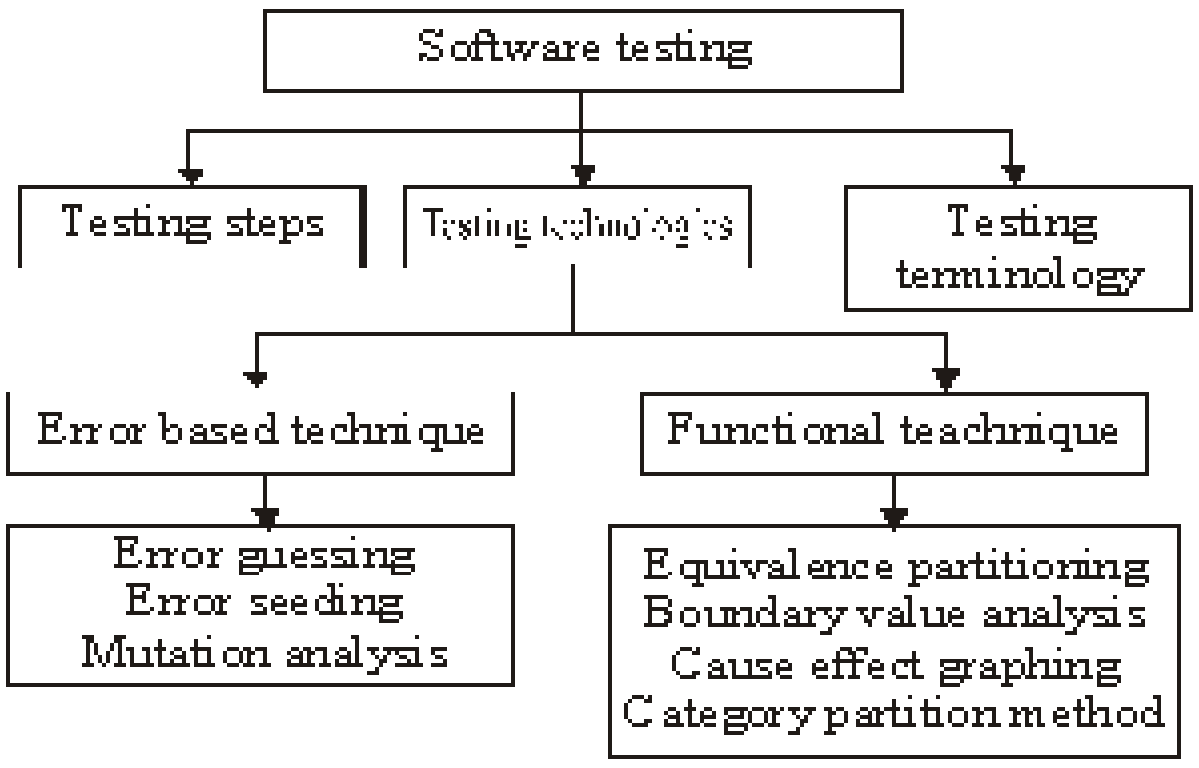

Figure 1: Representation of Classes

The class testing terminology relates to the class oop paradigm by the property isAppliedin. The class oop paradigm in turn relates with programming language via utilizes property and the class testing techniques links advanced concepts by are Automated By property. These properties have characteristics like antisymmetric and irreflexive.

The properties can be edited using the properties tab selecting either object properties or datatype properties. Annotations can also be added to the properties in order to describe about it. To create an object property switch to object properties tab, use the add object property button, this creates a new object property.

### *3.3 Open World Assumption*

The assumption is made by description logic, this denotes a lack of knowledge. The consequence is that if two classes testing phases and testing techniques are not defined as disjoint then it can have the individuals in common. The disjointness in classes plays a vital role in each of the class description. Creating a class and making it complement to another class is done here.

Reasoning capabilities are exploited to detect logical inconsistencies within the ontology. The error has been occurred while setting characteristics, asymmetric and reflexive to a same property. The consistency checks can help developer in an adequate manner while constructing the ontologies.

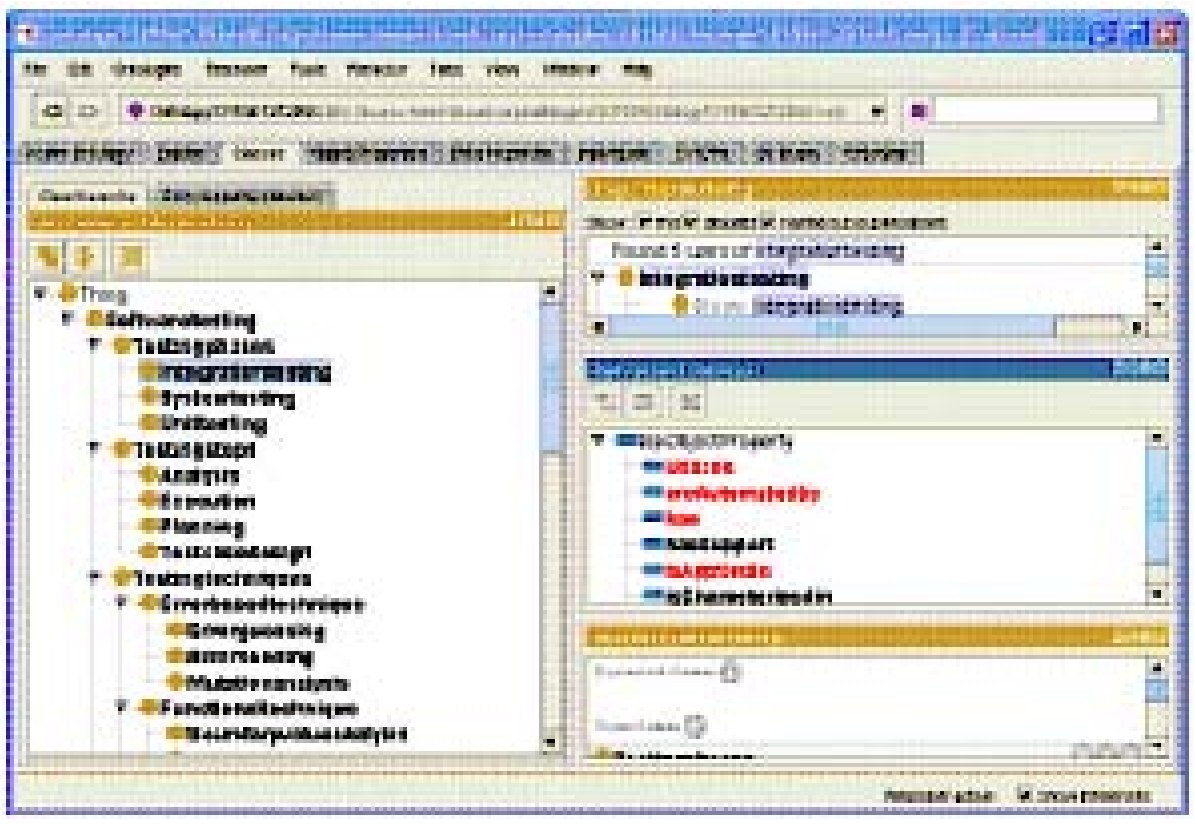

Figure 2: Class View

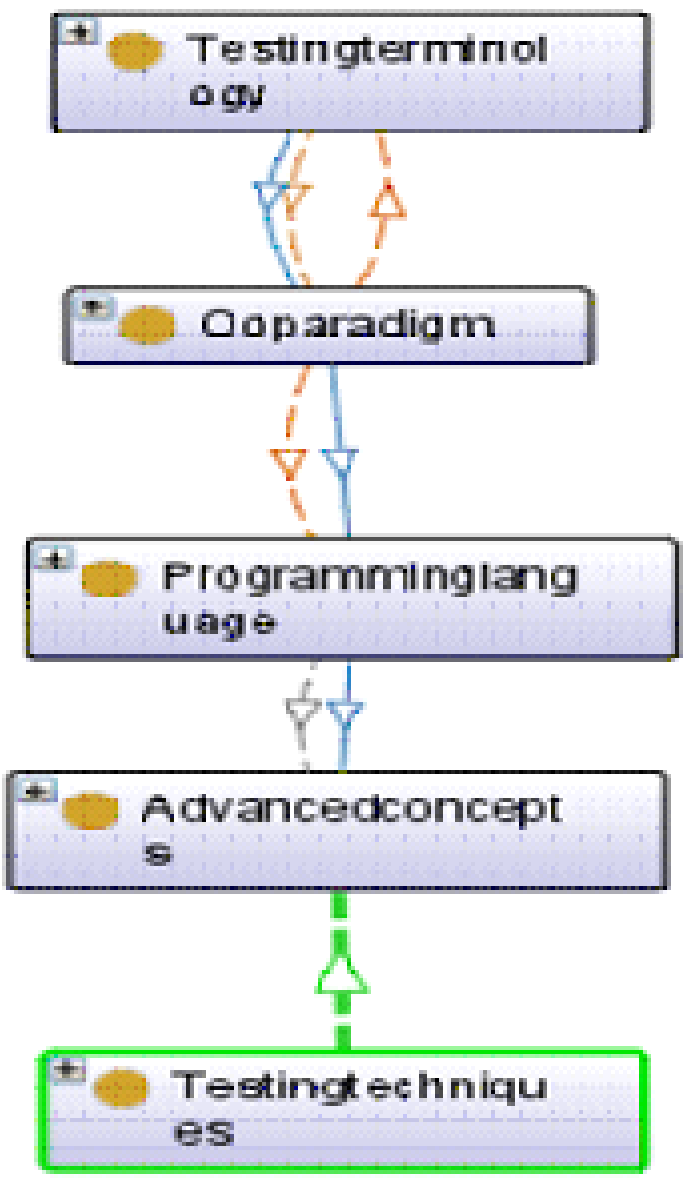

Figure 3: Property Relations in Classes

The important issue with reasoners is that OWL is not able to handle full expressivity. The specification distinguishes between OWL Full and OWL DL to indicate tractable language elements to reasoners. Ontologies which use metaclasses which is a OWL Full element cannot be classified. The conversion of OWL Full to OWL DL can be made using the classifier. Complete OWL Full syntax is not supported by protégé.

### 3.4 OWL/XML Rendering

The structure of any expression in RDF consists of triples, each consisting of a subject, a predicate and an object. A set of each triples is called an RDF graph. This can be illustrated using the node and arc diagram, in which each triple is represented as a node-arc-node link. In order to avoid conversion between different description languages, ontology needs a common language to express. XML has been used for this purpose since it has standards on data exchange. OWL ontology is most commonly serialized using OWL/XML syntax. The OWL/XML schemas are represented in figure 4.

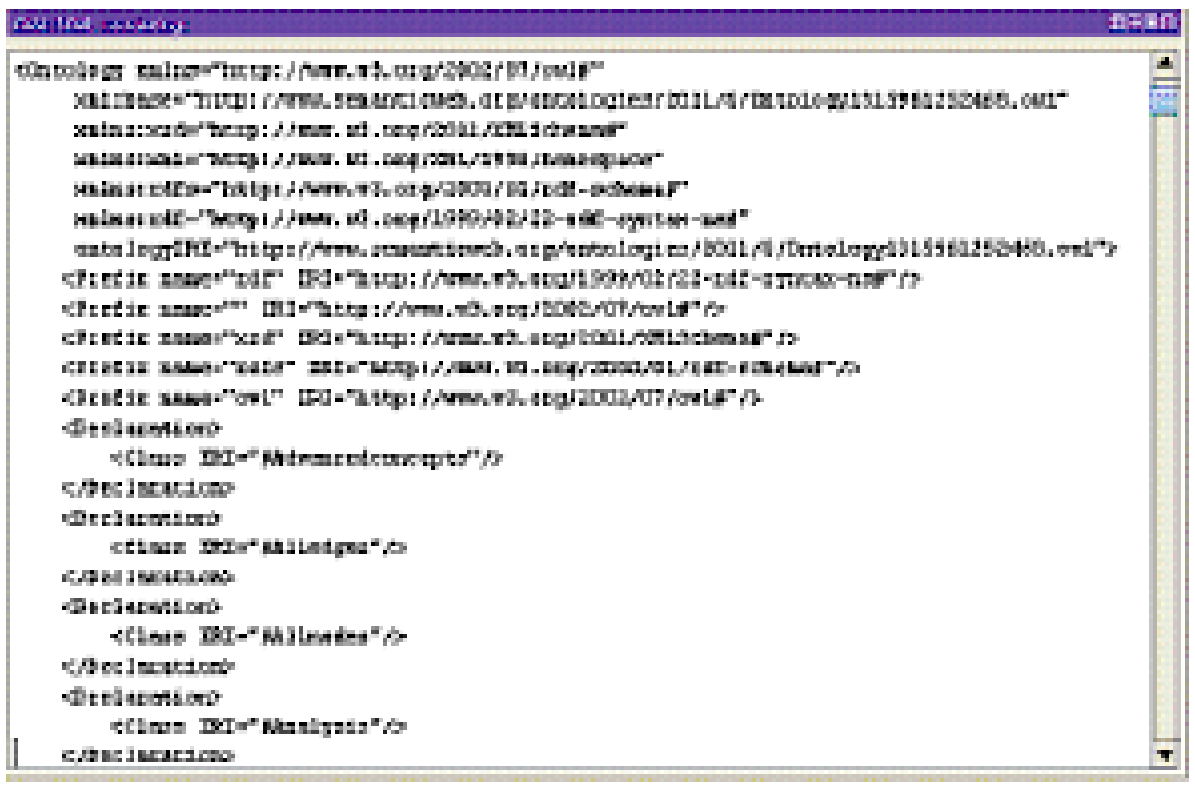

Figure 4: OWL/XML Schemas

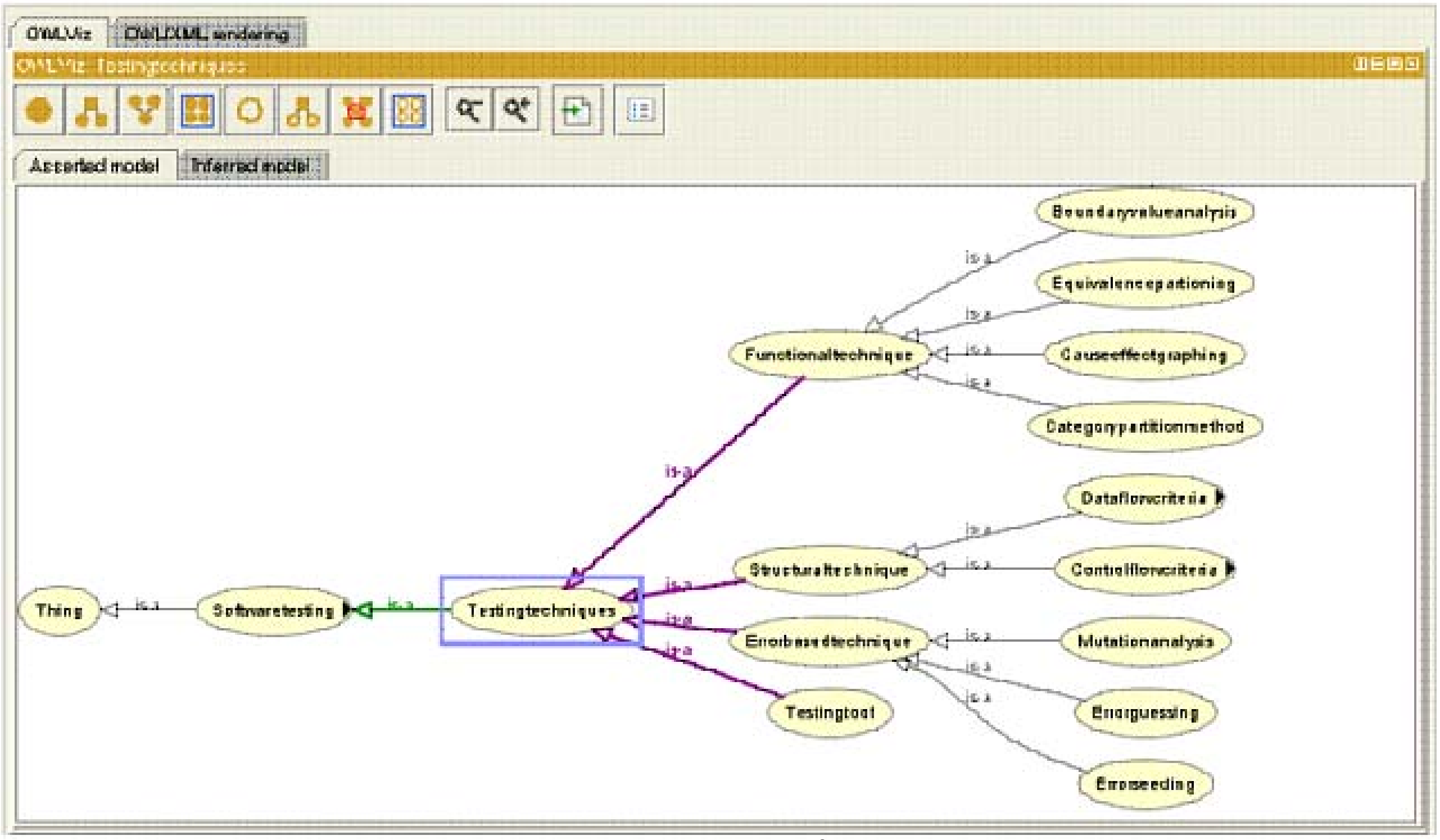

Figure 5: OWL Viz view of Classes

## 4. CONCLUSION AND FUTURE WORK

This paper described a framework of an ontology construction for integrated teaching of programming foundations with testing. Under this construction of framework of ontology, the programmers get depth knowledge about the application of testing concepts along with programming. This helps the software industry to train their freshers in a perfect manner.

The future work of this paper is to merge different ontologies using the same technique.